\documentclass[sigconf]{acmart}
\AtBeginDocument{%
  }

\setcopyright{acmlicensed}
\copyrightyear{2018}
\acmYear{2018}
\acmDOI{XXXXXXX.XXXXXXX}
\acmConference[Conference acronym 'XX]{Make sure to enter the correct
  conference title from your rights confirmation email}{June 03--05,
  2018}{Woodstock, NY}
\acmISBN{978-1-4503-XXXX-X/2018/06}

\usepackage{booktabs}
\usepackage{multirow}
\usepackage{makecell}
  
\usepackage{amssymb}
\usepackage{graphicx}
\usepackage{pifont}
\usepackage[table]{xcolor}
\usepackage{nicematrix}
\usepackage{svg}
\usepackage{tikz}
\usepackage{amssymb}
\usepackage{amsmath}
\usetikzlibrary{fit}

\usepackage[most]{tcolorbox}
\usepackage{fancyhdr}
\usepackage{float}
\renewcommand\footnotetextcopyrightpermission[1]{}
\usetikzlibrary{calc}

\usepackage{cleveref}

\acmSubmissionID{346}

\begin{document}


\title{SDesc3D: Towards Layout-Aware 3D Indoor Scene Generation from Short Descriptions}

\author{Jie Feng}
\email{jiefeng0109@163.com}
\affiliation{%
  \institution{Xidian University}
  \country{China}
}

\author{Jiawei Shen}
\email{jiaweishen@stu.xidian.edu.cn}
\affiliation{%
  \institution{Xidian University}
  \country{China}
}

\author{Junjia Huang}
\email{huangjj77@mail2.sys.edu.cn}
\affiliation{%
  \institution{Sun Yat-sen University}
  \country{China}
}

\author{Junpeng Zhang}
\email{junpengzhang@xidian.edu.cn}
\affiliation{%
  \institution{Xidian University}
  \country{China}
}

\author{Mingtao Feng}
\email{mintfeng@hnu.edu.cn}
\affiliation{%
  \institution{Xidian University}
  \country{China}
}

\author{Weisheng Dong}
\email{wsdong@mail.xidian.edu.cn}
\affiliation{%
  \institution{Xidian University}
  \country{China}
}

\author{Guanbin Li}
\authornote{Corresponding author.}
\email{liguanbin@mail.sysu.edu.cn}
\affiliation{%
  \institution{Sun Yat-sen University}
  \country{China}
}

\renewcommand{\shortauthors}{Trovato et al.}

\begin{abstract}
3D indoor scene generation conditioned on short textual descriptions provides a promising avenue for interactive 3D environment construction without the need for labor-intensive layout specification.  
Despite recent progress in text-conditioned 3D scene generation, existing works suffer from poor physical plausibility and insufficient detail richness in such semantic condensation cases, largely due to their reliance on explicit semantic cues about compositional objects and their spatial relationships. 
This limitation highlights the need for enhanced 3D reasoning capabilities, particularly in terms of prior integration and spatial anchoring.
Motivated by this, we propose SDesc3D, a short-text conditioned 3D indoor scene generation framework, that leverages multi-view structural priors and regional functionality implications to enable 3D layout reasoning under sparse textual guidance.
Specifically, we introduce a Multi-view scene prior augmentation that enriches underspecified textual inputs with aggregated multi-view structural knowledge, shifting from inaccessible semantic relation cues to multi-view relational prior aggregation. 
Building on this, we design a Functionality-aware layout grounding, employing regional functionality grounding for implicit spatial anchors and conducting hierarchical layout reasoning to enhance scene organization and semantic plausibility.
Furthermore, an Iterative reflection-rectification scheme is employed for progressive structural plausibility refinement via self-rectification.
Extensive experiments show that our method outperforms existing approaches on short-text conditioned 3D indoor scene generation.
Code will be publicly available.
\end{abstract}



\begin{teaserfigure}
  \centering
  \includegraphics[width=1.0\textwidth]{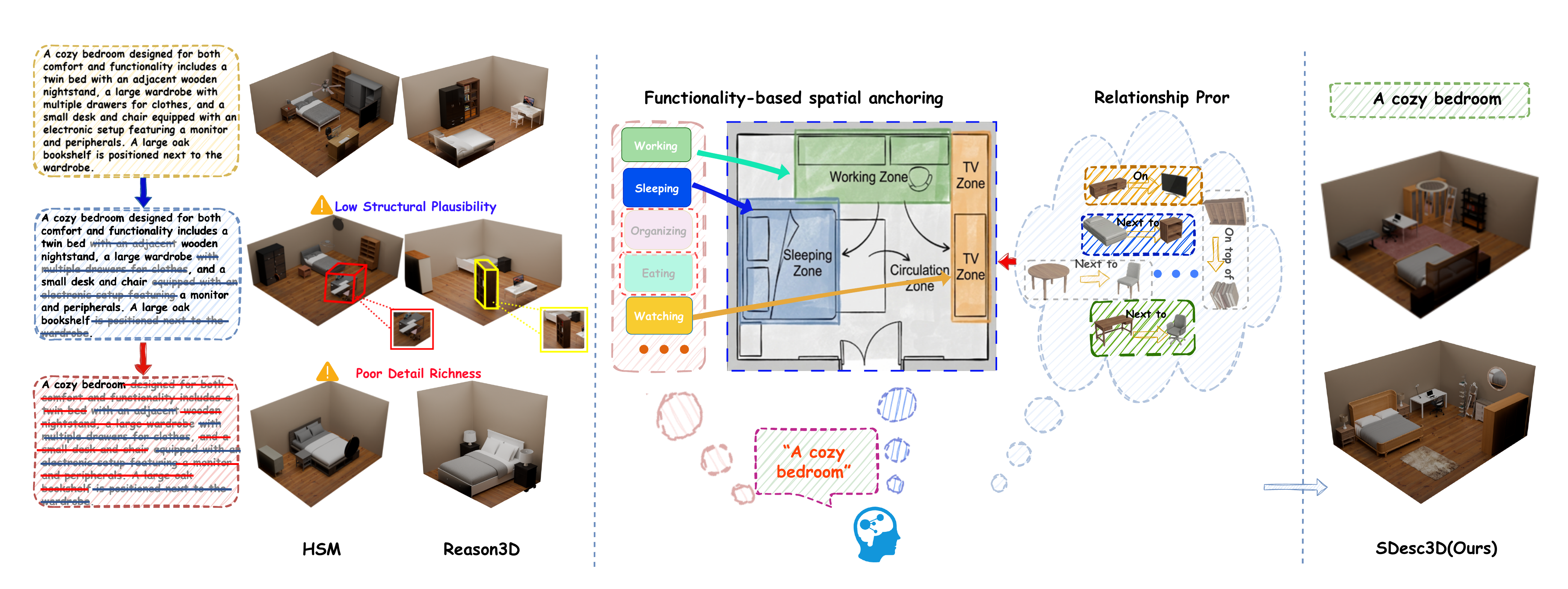}
  \caption{Short descriptions condense semantics, making 3D indoor scene generation challenging in terms of physical plausibility and detail preservation. 
  To address this, SDesc3D leverages object coherence priors and functionality-based spatial anchoring, achieving superior performance under such a challenging condition.}
  \label{fig:teaser}
\end{teaserfigure}


\maketitle

\section{Introduction}

Driven by the increasing demand for immersive interactions, 3D indoor scene generation has emerged as a cornerstone across diverse fields, from virtual content creation to embodied AI research, thanks to its ability to efficiently create interactive and scalable 3D environments while significantly reducing the manual effort required for scene modeling.
In particular, recent advances in text-conditioned 3D scene generation offer great potential by enabling users to convert natural language descriptions into detailed 3D layouts and scenes.
Pioneering works in this field have explored autoregressive\cite{maantmm2024} and diffusion\cite{diffuscenecvpr2024} models to enforce semantic coherence between textual descriptions and generated scenes, however, the generated scene diveristy is significantly constrained by the limited availability of large-scale 3D indoor scene datasets.
With the emergence of large language models (LLMs), LLM-guided 3D scene generation has emerged as a new paradigm, which expands scene diversity by leveraging common-sense knowledge from text for reasoning about spatial layouts \cite{hsm3dv2026, t2scene-lrmaaai2026}.

Generating 3D indoor scenes from short descriptions, such as "\textit{a cozy bedroom}", provides a promising avenue for interactive 3D environment construction without the need for labor-intensive layout specification.  
Despite recent advancements in LLM-guided 3D scene generation, existing approaches fail to handle such a challenging case with intense semantic condensation, due to their reliance on detailed scene descriptions that explicitly convey the semantic concepts of compositional elements as well as their corresponding spatial relationships.
As illustrated in Fig.~\ref{fig:teaser}, when spatial relationships are omitted from a detailed scene description, the generated scenes often exhibit structurally implausible or physically invalid object arrangements.  
Furthermore, if details about the compositional elements are also omitted, the richness and fidelity of the generated scenes deteriorate noticeably.
This highlights the critical gap in semantic richness and physical plausibility between semantic condensation in short textual description and the requirements for faithful 3D indoor scene generation.

Compensating for the lack of explicit semantic guidance on object relationships plays a crucial role in short-text conditioned 3D indoor scene generation.
To achieve this, RoomPlanner \cite{roomplanner2025} explores LLM-based description augmentation to expand short scene descriptions into executable and detailed descriptions, however, its performance remains limited by the reasoning capabilities of LLMs.
In contrast, Scenethesis \cite{scenethesisiclr2026} utilizes an image generation model to extracts scene layouts from the generated images conditioned on augmented descriptions, thereby improving realism and physical plausibility through structural knowledge acquired from large-scale pretrained image generation models.
Despite the potential of integrating auxiliary knowledge, using a single-view generated images remain limited in capturing the full 3D object arrangements, as view perspective, partial occlusion and fully missing object visibility can obscure crucial spatial relationships.
This limitation motivates the use of multi-view priors to bridge the semantic gap arising from inaccessible semantic guidance.

In this paper, we propose SDesc3D, a short-text conditioned 3D indoor scene generation framework, that leverages multi-view structural priors and regional functionality implications to enable 3D layout reasoning under sparse textual guidance.
First, we propose Multi-view scene prior augmentation (MSPA) to enriches short textual descriptions with priors aggregated from external multi-view scene resources.
Our MSPA constructs a semantic scene prior memory by encoding multi-view indoor scenes into relational representations. 
For a given short scene description, relevant priors are retrieved to augment the description, providing generalized guidance on plausible spatial relationships for guiding subsequent 3D layout reasoning.
With augmented scene description provides cues on possible object compositions and their coherence, plausible spatial arrangements remains unconstrained, due to the lack of spatial anchoring.
To this end, we highligt that short scene description and augmented priors provide valuable implications about the regional functionalities of the targeted scene, offering valuable cues on scene structure and organization at a coarser granularity, as illustrated in \Cref{fig:teaser}.
Building on this insight, we propose Functionality-aware Layout Grounding (FLG), where functionality zones are inferred by aggregating objects and leveraged as spatial anchors for reasoning about object-wise spatial arrangements. 
Furthermore, to maintain physical plausibility, we introduce an Iterative reflection-rectification (IRR) scheme that progressively refines the layout through a multi-stage and feedback-driven process.

Our main contributions are summarized as follows:
\begin{itemize}
    \item We propose Multi-view scene prior augmentation to enrich short descriptions with aggregated multi-view structural knowledge, shifting guidance from inaccessible semantic relation cues to multi-view relational prior aggregation.

    \item Our Functionality-aware layout grounding constructs a hierarchical layout reasoning process that leveraging regional functionalities for implicit spatial anchoring, leading to improved scene organization and semantic plausibility.

    \item An Iterative reflection-rectification scheme is designed for progressively layout refinement by leveraging feedback from each refinement stage to improve physical plausibility while presenting over-deterministic corrections via continous re-evaluating.

    \item Extensive experiments show that our method outperforms existing approaches on short-text conditioned 3D indoor scene generation, with enhanced physicial plausibility and improved detail richness.
\end{itemize}

\section{Related Work}

\begin{figure*}[t]  
  \centering
  \includegraphics[width=\textwidth]{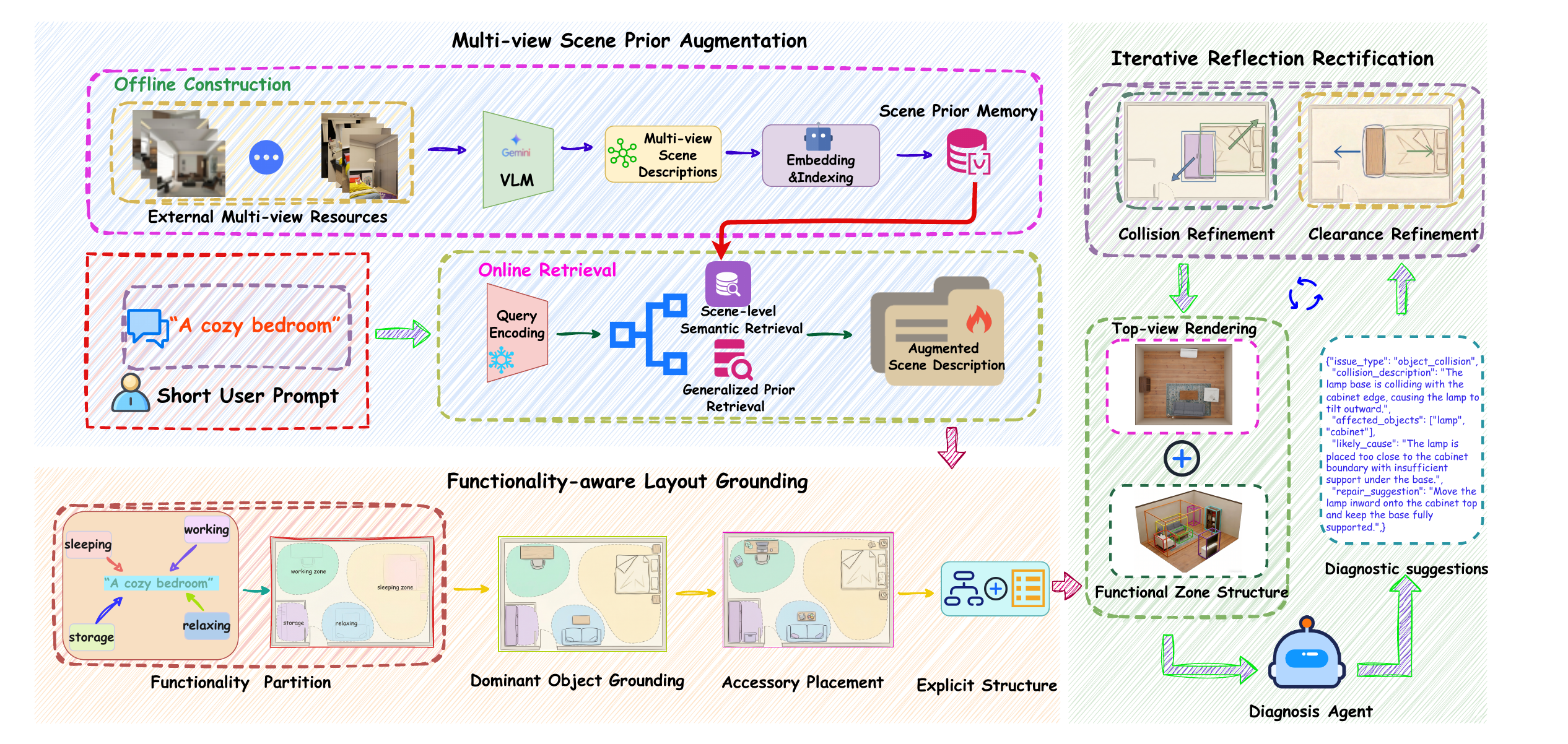}
  \caption{Overview of our SDesc3D Framework. Given a short user descriptions, SDesc3D first performs Multi-view Scene Prior Augmentation to retrieve scene priors for sparse semantic completion. Next, Functionality-aware Layout Grounding leverages regional functionality implications to reason about a hierarchical layout in a coarse-fine manner. Finally, Iterative reflection-rectification is adopted to iteratively suppress residual physical and structural errors.}
  \label{fig:overview}
  \Description{Overview of Our Framework.}
\end{figure*}

\subsection{Text-conditioned 3D Indoor Scene Generation}

Early 3D indoor scene generation mainly followed a data-driven layout synthesis paradigm~\cite{fastpriortvcg2022,graphmatchgm2016,grainstog2019}, where plausible indoor environments were produced by learning object co-occurrence patterns and spatial priors from scene datasets.
While effective under restricted data distributions, such methods offered only limited support for text-conditioned generation~\cite{changemnlp2014}, as scene composition was driven primarily by dataset regularities rather than explicit linguistic intent.

With the development of large language models (LLMs) and vision-language models (VLMs), text-conditioned 3D indoor scene generation has evolved into a more open paradigm involving language understanding, asset retrieval, layout planning, and constraint-aware optimization. 
Representative methods include LayoutGPT~\cite{layoutgptnips2023}, which maps natural language to layout representations through in-context learning; 
Holodeck~\cite{holodeckcvpr2024}, which integrates language understanding, asset retrieval, and spatial constraint optimization into an open-vocabulary generation pipeline; 
LayoutVLM~\cite{layoutvlmcvpr2025}, which combines VLMs with differentiable optimization to improve semantic consistency and physical plausibility; 
and Reason3D~\cite{t2scene-lrmaaai2026} , which unifies scene planning, object retrieval, and constraint-aware placement within a stronger reasoning framework. 
Despite these advances, most existing methods~\cite{commonscenesnips2023,anyhomeeccv2024,idesigneccv2024,openuniverse2024} still rely on the same implicit premise: the input text already specifies the key structural cues required for generation, including candidate objects, spatial relations, and other layout-relevant constraints. 
In contrast, this work focuses on brighing the critical gap in semantic richness and physical plausibility between semantic condensation in short textual description and the requirements for faithful 3D indoor scene generation.

\subsection{Hierarchical Reasoning for Progressive Generation}

Prior work~\cite{diffuscenecvpr2024,physcenecvpr2024,fireplacecvpr2025,gltreecvpr2025,hierovocabaaai2025,scenelanguagecvpr2025}  on controllable 3D scene generationhas progressively moved from implicit scene modeling toward explicit layout representations.
Early methods\cite{3dslncvpr2020,instructsceneiclr2024,graphcanvas2024} were largely formulated at the object and relation levels, using scene graphs, relation graphs, or programmatic structures to encode object attributes, inter-object relations, and local spatial constraints, thereby improving structural interpretability and object-level controllability. 
Imaginarium~\cite{imaginariumTOG2025} further extends this line by introducing scene-graph optimization after vision-guided 3D layout recovery, strengthening post-generation structural correction.

To improve structural organization in complex indoor scenes, subsequent works introduces hierarchical and multi-scale modeling that progressively unfolds scene generation in a coarse-to-fine manner. 
Architect~\cite{architectnips2024} adopts hierarchical 2D inpainting for progressive large-scale 3D scene generation, 
HSM~\cite{hsm3dv2026} models cross-scale support regions and object compositions through hierarchical scene motifs, 
and HiScene~\cite{hiscenemm2025} treats a scene as a hierarchy of decomposable entities to preserve consistency between global structure and local instances. 
Compared with purely local relation-based formulations, these methods strengthen multi-level structural organization in progressive generation. However, their hierarchical mechanisms remain largely grounded in support relations, local compositional patterns, or structural priors, and are therefore better suited to organizing existing layout elements than to revealing the intermediate structure required for layout reasoning under sparse descriptions. 
Under short scene descriptions, stable spatial anchors are often absent, while the implicit functionality cues embedded in text remain underexploited.

\subsection{Post-reasoning Layout Optimization}

After an initial layout is generated, residual geometric conflicts and local inconsistencies may still remain, making post-reasoning layout optimization necessary for improving physical plausibility.
Early approaches~\cite{planittog2019,humancentriccvpr2018,hybridreptog2020,placementprog2025} mainly treat this stage as local post-hoc repair, resolving explicit violations such as object penetration, boundary overflow and unstable placement through collision handling, constraint solving, or rule-based geometric adjustment.
For example, PhyScene~\cite{physcenecvpr2024} improves layout feasibility with physical and interaction constraints, while FirePlace~\cite{fireplacecvpr2025} further rectifies object placement through geometric constraint construction and solving.
Despite their effectiveness, these methods still focus on exposed local defects in a single shot rather than optimizing the intermediate layout for progressive refinement.

More recent methods~\cite{directlayoutneurips2025,scenelcm2025,scenecrafticml2024} introduce feedback and tool-assisted refinement.
SceneWeaver~\cite{sceneweavernips2025} integrates iterative scene synthesis with a LLM-based refinement tool, while DisCo-Layout~\cite{discolayout2025} decouples semantic and physical refinement via multi-agent collaboration.
However, their refinement remains largely violation-driven, with correction solely applied to defects revealed at current step.
As a result, existing one-shot and iterative adjustment may improve local feasibility, yet still compromise scene-level structure and semantic consistency.
In this work, we formulate post-reasoning layout optimization as a multi-step reflection-rectification process while preserving the global scene structure.

\section{Methodology}

We begin by formulating the task of 3D indoor scene generation conditioned on a short description.
Given a short scene description $d$, such as "\textit{a cozy bedroom}", which only provides the scene type, the goal is to generate a 3D indoor scene $s$ without relying on explicit description about the internal compositional object instances and their spatial relationships. 
This necessitates strong reasoning abilities in inferring compositional objects and their spatial arrangements, while ensuring alignment with physical and functional plausibility under minimal guidance from a condensed description.

To achieve this goal, we introduce SDesc3D, a short-text conditioned 3D indoor scene generation framework, that leverages multi-view structural priors and regional functionality implications to enable 3D layout reasoning under sparse textual guidance.
As illustrated in \Cref{fig:overview}, our SDesc3D consists of three core components: Multi-view Scene Prior Augmentation, Functionality-aware Layout Grounding and Iterative Reflection-Rectification. 
Given a short scene description $d$, multi-view scene prior augmentation produces an augmented scene description with assistance of multi-view scene priors (\Cref{subsec:prior_augmentation}).
Based on the augmented scene description $d_{\text{aug}}$, functionality-aware layout grounding is then employed to construct an initial layout $\mathcal{G}^{0}$ in a hierarchical manner (\Cref{subsec:layout_grounding}). 
Subsequently, the obtained functionality-aware layout $\mathcal{G}^{0}$ is passed to our Iterative Reflection-Rectification for progressive refinement (\Cref{subsec:irr}).

\subsection{Multi-view Scene Prior Augmentation}
\label{subsec:prior_augmentation}

A core obstacle in 3D indoor scene generation conditioned on short and under-specified descriptions is the absence of semantic cues regarding detailed composition and object relationships, necessitating the demands on auxiliary priors in 3D layout reasoning. 
Visual guidance has shown promising capability in providing scene priors \cite{scenethesisiclr2026}, however, it suffers from unsatisfactory scene understanding, due to partial occlusion and fully missing object visibility from a single-view perspective.
To mitigate this issue, we propose Multi-view scene prior augmentation (MSPA), which enriches short textual descriptions with priors aggregated from multi-view images.
Instead of extracting object arrangements directly from a generated image, MSPA constructs a semantic scene prior memory by encoding multi-view indoor scenes from external resources into relational representations.
Given a short scene description, relevant priors are retrieved based on semantic or generalized similarity and used to augment the description, providing generalized guidance on plausible spatial relationships for guiding subsequent 3D layout reasoning.

\textbf{Offline Multi-view Scene Prior Construction.}
To provide guidance on object relationships that is otherwise inaccessible, a multi-view scene prior memory is constructed from external scene resources.
Specifically, we collect multi-view indoor images from ScanNet~\cite{scannetcvpr2017} and SpatialGen~\cite{spatialgen3dv2026}.
For the $i$-th collected scene, we sample $N$ images from distinct perspectives and group them into a multi-view observation unit $\mathcal{V}_i$, providing a comprehensive view of object relationships in the scene.
Then we employ the vision-language model Gemini-3 Flash to jointly parse the observation unit $\mathcal{V}_i$ into three complementary types of textual descriptions,
\begin{equation}
\{\mathcal{I}_{\text{sum}}^{i}, \mathcal{I}_{\text{rel}}^{i}, \mathcal{I}_{\text{scale}}^{i}\}
=
\Phi_{\text{vlm}}(\mathcal{V}_i),
\end{equation}
where $\mathcal{I}_{\text{sum}}^{(i)}$ encodes a brief summary on the scene type, and $\mathcal{I}_{\text{rel}}^{(i)}$ contains detailed textual descriptions on object relationships. 
$\mathcal{I}_{\text{scale}}^{(i)}$ captures the relative object scales in the given scene.

Based on these parsed descriptions, each scene’s type summary, object relationships and scale information are organized into a structured scene entry $\tau_i$, which serves as the basic unit of the scene prior memory. 
Inspired by \cite{layoutgptnips2023}, we encode each structured scene entry in the structural programming language, for exploring the spatial reasoning capabilities of LLMs.
By traversing all collected multi-view scenes, a scene prior memory is constructed as
\begin{equation}
\mathcal{M}=\{(\tau_i,z_i)\}_{i=1}^{|\mathcal{M}|},
\end{equation}
where $z_i$ refers to the embedding of $i$-th entry's summary $\tau_i$ for indexing.
By this way, the constructed memory $\mathcal{M}$ stores both multi-view scene priors and their semantic embeddings, thus supporting efficient similarity-based retrieval from short scene descriptions in our next stage.

\textbf{Retrieval-based Online Prior Augmentation.}
Given a short scene description \(d\), we propose augmenting the condensed description with retrieved scene priors from the constructed scene prior memory $\mathcal{M}$.
To achieve this, we first define a scene-level semantic retrieval score based on the semantic similarity between the scene summary of each memory entry $z_i$ and the given short scene description $d$,
\begin{equation}
s_{ret}(d,\tau_i)=\cos(d, z_i),
\end{equation}
where we omit the explicit embedding operator applied to $d$ for ease of notation.
Then, the top-$K$ memory entries with highest retrieval scores are collected to construct an aggregated scene prior package for description augmentation, denoted as $\mathcal{P}_{aug}$.
This package provides rich semantic guidance on compositional objects and object arrangements, drawing from the memory entries with similar scene-level semantic similarities.

While scene-level semantic similarity-based retrieval often yields effective priors, it can also aggregate misleading priors from unrelated scenes due to limited scene-type coverage in $\mathcal{M}$. 
When no sufficient semantic similarity is found, \textit{i.e.}, when $\max_{i=1}^{|\mathcal{M}|} s_{ret}(d,\tau_i) \le \theta_{ret}$ with $\theta_{ret}$ being a pre-set threshold, we propose an alternative retrieval scheme to aggregate generalized object relationships from the prior memory instead.
To this end, we define a generalized prior retrieval score based on the Best matching-25 (BM25) ranking function \cite{robertson1995okapi} as
\begin{equation}
s_{ret}(d,\tau_i)=\text{BM25}(d, \tau_i),
\end{equation}
where the short description $d$ is temporarily enriched with potential objects and used to quantify its relationship with $\tau_i$ based on the term occurrences. 
By adopting this retrieval score, priors with possibly similar compositional objects are aggregated to provide generalized knowledge on their spatial relationships.
Similarly, a prior package $\mathcal{P}_{aug}$ is constructed by selecting memory entries with top-K retrieval scores.

Finally, the augmented scene description is constructed by combining the given short description with the retrieved prior package, 
\begin{equation}
d_{\text{aug}}=(d,\mathcal{P}_{\text{aug}}),
\end{equation}
bridging the semantic gap arising from inaccessible semantic guidance by enhanced scene priors on plausible layouts as well as generalized object spatial relationships.

\begin{table*}[t]
\centering
\caption{Quantitative comparison under the short-text 3D indoor scene generation setting. Compared with HSM and Reason3D, our method achieves improved overall performance.}
\vspace{-10pt} 
\label{tab:ablation_compare}
\setlength{\tabcolsep}{3.5pt}
\footnotesize
\begin{tabular}{@{}lcc*{8}{cc}@{}}
\toprule
\multirow{2}{*}{\textbf{Method}}
& \multirow{2}{*}{\makecell{\textbf{Collision (\%)$\downarrow$}}}
& \multirow{2}{*}{\makecell{\textbf{OOB (\%)$\downarrow$}}}
& \multicolumn{2}{c}{\makecell{\textbf{OP$\uparrow$}}}
& \multicolumn{2}{c}{\makecell{\textbf{AO$\uparrow$}}}
& \multicolumn{2}{c}{\makecell{\textbf{DR$\uparrow$}}}
& \multicolumn{2}{c}{\makecell{\textbf{ZO$\uparrow$}}}
& \multicolumn{2}{c}{\makecell{\textbf{CR$\uparrow$}}}
& \multicolumn{2}{c}{\makecell{\textbf{FC$\uparrow$}}}
& \multicolumn{2}{c}{\makecell{\textbf{AI-Avg$\uparrow$}}} \\
\cmidrule(lr){4-5}
\cmidrule(lr){6-7}
\cmidrule(lr){8-9}
\cmidrule(lr){10-11}
\cmidrule(lr){12-13}
\cmidrule(lr){14-15}
\cmidrule(lr){16-17}
\cmidrule(lr){18-19}
& & 
& \makecell{\scriptsize Gemini\\[-1pt]\scriptsize 3-flash}
& \makecell{\scriptsize GPT\\[-1pt]\scriptsize 5.4}
& \makecell{\scriptsize Gemini\\[-1pt]\scriptsize 3-flash}
& \makecell{\scriptsize GPT\\[-1pt]\scriptsize 5.4}
& \makecell{\scriptsize Gemini\\[-1pt]\scriptsize 3-flash}
& \makecell{\scriptsize GPT\\[-1pt]\scriptsize 5.4}
& \makecell{\scriptsize Gemini\\[-1pt]\scriptsize 3-flash}
& \makecell{\scriptsize GPT\\[-1pt]\scriptsize 5.4}
& \makecell{\scriptsize Gemini\\[-1pt]\scriptsize 3-flash}
& \makecell{\scriptsize GPT\\[-1pt]\scriptsize 5.4}
& \makecell{\scriptsize Gemini\\[-1pt]\scriptsize 3-flash}
& \makecell{\scriptsize GPT\\[-1pt]\scriptsize 5.4}
& \makecell{\scriptsize Gemini\\[-1pt]\scriptsize 3-flash}
& \makecell{\scriptsize GPT\\[-1pt]\scriptsize 5.4} \\
\midrule

HSM (3DV 2026)~\cite{hsm3dv2026}
&\makecell{9.36\%}    & \makecell{8.67\%}
&\textbf{9.31}  &8.83
&\textbf{9.34}  &8.96
&7.83  &7.47
&8.32  &8.61
&7.73  &7.48
&8.51  &7.96
&8.51  &8.22  \\

Reason3D (AAAI 2026)~\cite{t2scene-lrmaaai2026}
&\makecell{18.62\%}    & \makecell{12.55\%}
&8.87  &8.32 
&9.32  &\textbf{8.99} 
&7.87  &7.94 
&8.81  &8.63 
&7.69  &7.43 
&8.07  &7.86 
&8.44  &8.20  \\

Ours
&\textbf{\makecell{5.36\%}}  &\textbf{\makecell{7.70\%}}  
&9.15  &\textbf{8.89} 
&9.26  &8.71 
&\textbf{8.21}  &\textbf{8.24} 
&\textbf{9.67}  &\textbf{9.04} 
&\textbf{9.02}  &\textbf{8.56} 
&\textbf{9.03}  &\textbf{9.01} 
&\textbf{9.06}  &\textbf{8.74}   \\

\bottomrule
\end{tabular}
\end{table*}

\subsection{Functionality-aware Layout Grounding}
\label{subsec:layout_grounding}

While our MSPA enriches the scene description with detailed semantic priors about compositional object instances and their spatial relationships, layout reasoning remains a challenging problem, due to the absence of spatial anchors in textual descriptions.
Beyond these explicit spatial relationships, the augmented description also provides valuable implications about the regional functionalities of the targeted scene, offering valuable cues on scene structure and organization at a coarser granularity.
Building on this insight, we propose Functionality-aware Layout Grounding (FLG), where functionality zones are inferred by aggregating objects and leveraged as spatial anchors for reasoning about object-wise spatial arrangements. 
Specifically, we partition the scene into functionality zones at a coarse semantic granularity. 
At a finer level, each zone is composed of one or more dominant objects, along with optional accessories, capturing both the core and supplementary elements that characterize the zone’s intended function.

To obtain such a hierarchical layout conditioned on the augmented description $d_{\text{aug}}$, we start with extracting functionality cues from the augmented description $d_{\text{aug}}$ and grounding extracted functionalities into spatial regions via an LLM agent.
This formulates a process as
\begin{equation}
    (\mathcal{Z}, \mathcal{F}) = \Phi_{\text{layout}}(d_{\text{aug}}),
\end{equation}
where $\mathcal{Z}$ refers to the set  of functionality zones, with each zone $\mathcal{Z}_i$ associated with a functionality type $\mathcal{F}_i$.

Then, starting from an empty layout $\mathcal{O}^{0} = \emptyset$, dominant objects are progressively arranged by traversing all functionality zones, 
\begin{equation}
\mathcal{O}^{t} = \Phi_{\text{layout}}(d_{\text{aug}}, \mathcal{Z}_t, \mathcal{F}_i, C(\mathcal{Z}_t) |\mathcal{O}^{t-1}), \quad \forall t \in \{1, \cdots, |\mathcal{Z}| \},
\end{equation}
where $\mathcal{Z}_t$ serves as a spatial anchor for local arrangement of dominating objects, and $C(\mathcal{Z}_t)$ refers to boundary constraint derived from $\mathcal{Z}_t$.
Since the functionality partition in real scenes does not strictly adhere to rigid boundaries, the constraint $C(\mathcal{Z}_t)$ is defined in a relaxed manner to allow flexible object arrangement across adjacent functionality zones.
For each functionality zone $\mathcal{Z}_t$, the constraint $C(\mathcal{Z}_t)$ is composed of two spatially expanded boundaries with decreasing priorities,
\begin{equation}
C(\mathcal{Z}_t) = \{C_{\text{buf}}(\mathcal{Z}_t), C_{\text{int}}(\mathcal{Z}_t)\},
\quad \text{s.t.} \ C_{\text{buf}}(\mathcal{Z}_t) \subset C_{\text{int}}(\mathcal{Z}_t),
\end{equation}
where $C_{\text{buf}}(\mathcal{Z}_t)$ creates an additional buffer for handling inter-object clearance.
The remaining $C_{\text{int}}(\mathcal{Z}_t)$ constructs an interactive space between adjacent functionality zones, and object arrangement is permitted with lower priority if the targeted space is unoccupied.

Subsequently, with the dominant object layout $\hat{\mathcal{O}} = \mathcal{O}^{|\mathcal{Z}|}$, we apply a finer-grained layout reasoning to arrange the accessories semantically linked with each dominant object, thus complementing its supporting functionality:
\begin{equation}
\mathcal{A}^t = \Phi_{\text{layout}}(d_{\text{aug}}, \hat{\mathcal{O}}_{t},C(\hat{\mathcal{O}}_t), S(\hat{\mathcal{O}}_t) | \mathcal{A}^{t-1}),
\quad \forall t \in \{1, \cdots, |\hat{\mathcal{O}}|\},
\end{equation}
where $\mathcal{A}^0 = \emptyset$, and an additional set of accessory arrangement criteria $S(\hat{\mathcal{O}}_t)$ is employed by following the identical settings from HSM \cite{hsm3dv2026}.
The final accessory layout is then denoted as $\hat{\mathcal{A}} = \mathcal{A}^{|\hat{\mathcal{O}}|}$.

Finally, a scene layout conditioned on the augmented scene description is constructed as 
\begin{equation}
\mathcal{G}^0 = \{\mathcal{Z}, \hat{\mathcal{O}}, \hat{\mathcal{A}}, \mathcal{H}, \mathcal{R}\}
\end{equation}
where $\mathcal{H}$ encodes the hierarchical relationships of the generated layout, and $\mathcal{R}$ captures the inter-object relationship at each semantic granularity.
By this functionality-aware and hierarchical reasoning design, the obtained layout demonstrates well-grounded functionality, enhanced spatial anchoring and improved scene richness.

\subsection{Iterative Reflection-Rectification}
\label{subsec:irr}

\begin{figure*}[t!]
  \centering
  \includegraphics[width=0.95\linewidth]{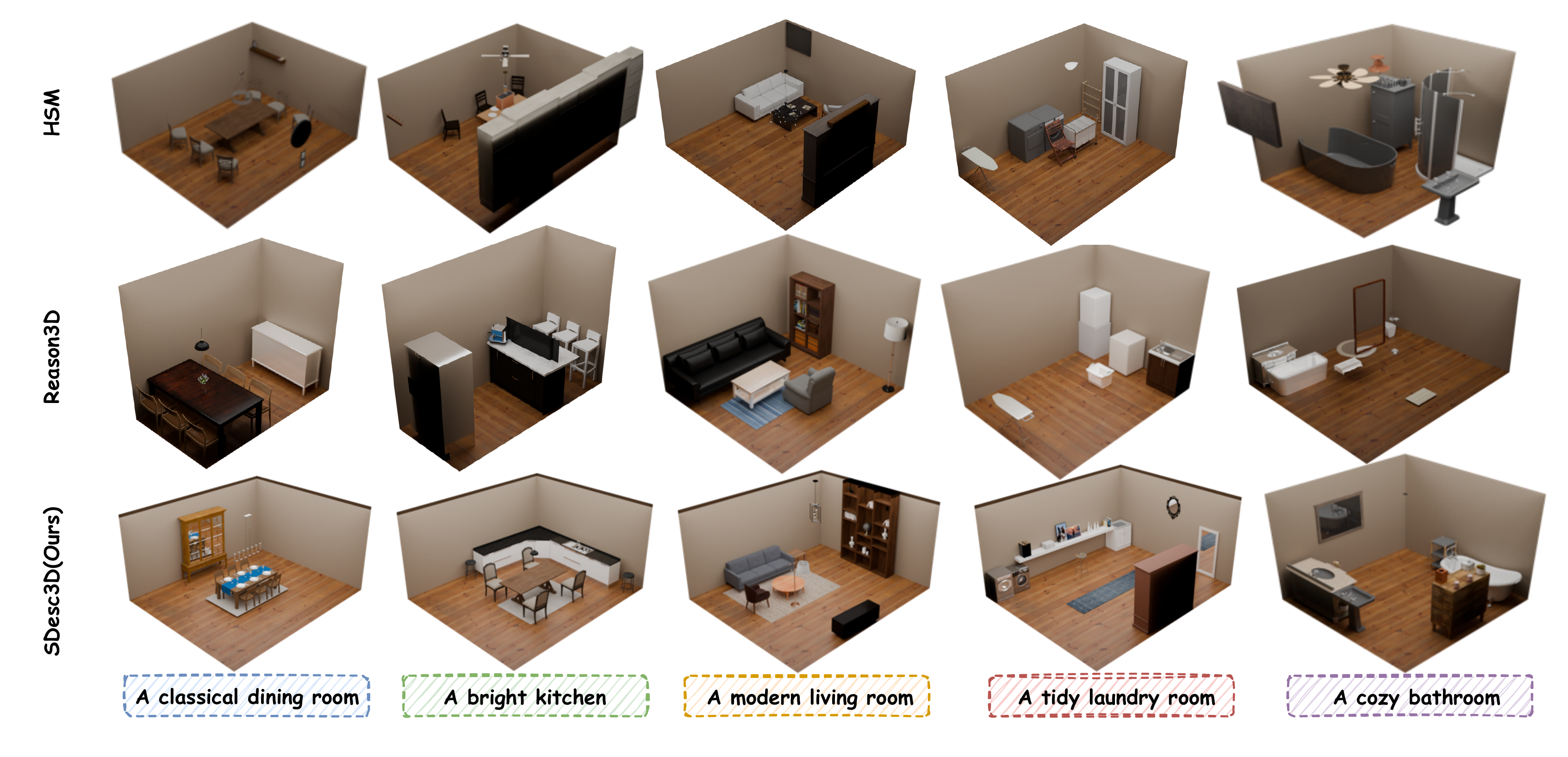}
  \caption{Qualitive comparison on the scenes generated on five different short descriptions. Our method achieves better overall scene quality than the compared approaches in terms of physical plausibility and detail richness.}
  \label{fig:Comparison of different methods}
\end{figure*}

Through our MSPA and FLG, the initial layout generated from a condensed scene descrition exhibits strong semantic plausibility and rich compositional objects, while it may not fully ensure physicial plausibilty, resulting in geometric violations such as object penetration, insufficient clearance and out-of-bound placement.
To maintain physical plausibility, existing approaches commonly employ a single-pass post-reasoning refinement to adjust object arrangements based on geometric criteria, which can introduce over-deterministic corrections and compromise scene integrity. 
In contrast, we introduce an Iterative reflection-rectification (IRR) scheme that progressively refines the layout through a multi-stage and feedback-driven process.
To achieve this, our IRR incorporates an LLM-guided diagnosis agent for identifying potential geometric violations and providing refinement suggestions at each intermediate layout, after which rectifications by a set of external tools is applied according for continous re-evaluting in the subsequent stage.

At each intermediate stage $t$, to complement the obtained layout $\mathcal{G}^{t-1}$ in structured programming language with auxiliary visual guidance, a top-down view image $\mathbf{I}^{t-1}$ is rendered, providing global contextual cues and spatial awareness that facilitate informed diagnose regarding structural violations. 
The intermediate layout $\mathcal{G}^{t-1}$, together with the rendered image $\mathbf{I}^{t-1}$, is then fed to our LLM-guided diagnosis agent $\Phi_{\text{diag}}(\cdots)$ for producing a textual diagnosis report,
\begin{equation}
\mathcal{D}^{t}=\Phi_{\text{diag}}(\mathcal{G}^{t-1}, \mathbf{I}^{t-1}, \mathcal{T}^{0:t-1}),
\end{equation}
where the refinement trace $\mathcal{T}^{0:t-1}$, containing the historical diagnoses and refinement feedbacks, is also incorporated to enable history-aware evaluation and stablize the refinement trajectory.

Since the obtained textual diagnosis alone cannot serve as a qualitive indicator, we define a layout penalty score by parsing the identified violations in $\mathcal{D}^{t}$ as
\begin{equation}
\label{eq:layout_penalty_score}
p_{\mathcal{D}^{t}}=\lambda_{pen} N_{pen}+\lambda_{clr} N_{clr}+\lambda_{oob} N_{oob},
\end{equation}
where $N_{pen}$, $N_{clr}$, and $N_{oob}$ denote the number of penetration pairs, object pairs with insufficient local clearance, and out-of-bound objects in $\mathcal{D}^{t}$, respectively.
The corresponding weighting factors, $\lambda_{\text{pen}}$, $\lambda_{\text{clr}}$ and $\lambda_{\text{oob}}$, control the relative importance of each violation type.

For an intermediate layout with a layout penalty score greater than a pre-set threshold, $p_{\mathcal{D}^{t}} \ge \theta_{p}$, a set of external rectification tools are applied to correct geometric violations.
In this work, we employ a clearance refinement tool to resolve improper distances among neighboring objects, based on the suggestions associated in the diagnosis report $\mathcal{D}^{t}$. 
At the same time, a collision refinement tool is adopted for handling the object penetration and out-of-bound issues.
The iterative refinement continues until either the layout penalty score falls below the threshold, $p_{\mathcal{D}^{t}} < \tau_{p}$, or the maximum number of refinement steps is reached, at which point the final layout is produced. 
Despite the simplicity of our adopted rectification tools, the effectiveness of our IRR is ensured, as it enables progressive error identification and correction across multiple refinement steps.
Additional details on our rectification tools are provided in the supplementary material.

\section{Experiment}

\subsection{Implementation and evaluation details}

For fair comparison, SDesc3D is evaluated under the same short-text setting, room-boundary constraints and shared 3D assets with the compared LLM-guided baselines.
All methods use a unified room scale configuration and a common object retrieval space constructed from Habitat Synthetic Scenes Dataset (HSSD-200)~\cite{hssdcvpr2024}, ensuring consistency in both asset sourcing and scene composition. 
For performance evaluation, we evaluate 50 test scenes and repeat each scene 10 times, reporting average results per-scene and over all scenes. 
For ablation study, we repeatedly evaluate the representative scene ``A cozy bedroom'' for 20 runs and report averaged metrics.

\textbf{Implementation details.}
Following \cite{t2scene-lrmaaai2026}, SDesc3D is instantiated with Gemini 3 Flash.
The multi-view prior memory is constructed from 144 reference scenes, each associated with 6 views. 
For each input short description, the top-3 most relevant priors are retrieved for augmentation.
The temperature is set to 0.3 for suggestion generation and 0.7 for all other LLM agents. 
For layout quality evaluation in \Cref{eq:layout_penalty_score}, we use a threshold $\theta_p=3.5$, with $\lambda_{\text{pen}}=0.3$, $\lambda_{\text{clr}}=0.2$ and $\lambda_{\text{oob}}=0.5$.
Additional details are provided in the supplementary material.

\textbf{Evaluation metrics.}
We evaluate the generated scenes from four complementary perspectives: physical plausibility, structural and semantic alignment, functional composition completeness, and scene richness. 
Collision (\%) and OOB (\%) measure the degree of inter-object collision and the proportion of out-of-bound objects, respectively. 
To assess whether short-text intent is effectively grounded into fine-grained scene structure, we further employ Gemini 3 Flash and GPT-5.4 as independent AI judges.
Semantic evaluation is organized into three categories: physical and structural plausibility, including Object Placement(OP) and Alignment and Order(AO); functional composition, including Functional Completeness(FC), Zone Organization(ZO), and Cross-zone Relation(CR); and scene richness, including Detail Richness(DR). 
The mean of all semantic scores is reported as AI-Avg. 
We further conduct a user study with 20 participants.
For each of the two paired score sets composed of AI-Avg and User-Avg on the same scenes, we report both $Pearson$ and $Spearman$ correlation coefficients to quantify the consistency between automatic evaluation and human judgment. 
Additional details on evaluation metrics are provided in the supplementary material.

\begin{table}[htbp]
\centering
\caption{User study of different methods under the short-text 3D indoor scene generation setting. Our method achieves the best results across all evaluated metrics.}
\vspace{-10pt} 
\label{tab:ablation_compare2}
\setlength{\tabcolsep}{3pt} 
\footnotesize
\begin{tabular}{lccccccccc} 
\toprule
\textbf{Method}

& \makecell{\textbf{OP}$\uparrow$}
& \makecell{\textbf{AO}$\uparrow$}
& \makecell{\textbf{DR}$\uparrow$}
& \makecell{\textbf{ZO}$\uparrow$}
& \makecell{\textbf{CR}$\uparrow$}
& \makecell{\textbf{FC}$\uparrow$}
& \makecell{\textbf{User-Avg}$\uparrow$} \\
\midrule
HSM (3DV 2026)~\cite{hsm3dv2026}        &7.26  &7.72  &7.46  &7.40  &7.52  &7.49  &7.48   \\
Reason3D (AAAI 2026)~\cite{t2scene-lrmaaai2026}  &7.15  &7.40  &7.36  &7.00  &6.94  &7.44  &7.22    \\
Ours                 &\textbf{8.64}   &\textbf{8.85}   &\textbf{8.69}   &\textbf{8.74}   &\textbf{8.77}   &\textbf{8.56}   &\textbf{8.71}      \\
\bottomrule
\end{tabular}
\end{table}

\subsection{Main Results}  
\textbf{Qualitative Comparison.} As shown in Fig.~\ref{fig:Comparison of different methods}, under short-text inputs, both HSM and Reason3D fail to consistently recover indoor layouts with sufficient completeness and clear organization. 
Reason3D largely remains at a coarse-grained generation level dominated by a few major furniture items, often resulting in missing auxiliary objects, insufficient functional support, and weak cross-region structure.
Although HSM is able to generate more objects, it still frequently produces loose object compositions, ambiguous functional boundaries, and local configurations that are not fully aligned with the intended scene semantics. 
In contrast, our method recovers more complete object sets from concise prompts and further establishes clearer functional partitioning and more coherent spatial organization, while maintaining better local arrangement stability and global layout consistency. 
This advantage comes from the joint effect of multi-view scene prior augmentation(MSPA), functionality-aware layout grounding(FLG) and iterative reflection-rectification(IRR), which together enable more effective object completion and layout structuring from sparse semantics, thereby substantially improving generation quality in the short-text setting.

\textbf{Quantitative Comparison.}Table~\ref{tab:ablation_compare} shows that our method achieves the best physical plausibility, reducing Collision and OOB to 5.36\% and 7.70\%, respectively, both substantially lower than those of HSM and Reason3D.
More importantly, under both AI judges, our method performs best on DR, ZO, CR, FC, and AI-Avg, with the most consistent gains observed on ZO, CR, and FC. 
These results show that our method not only recovers missing objects and fine-grained details from short-text inputs, but also more effectively establishes clear functional partitioning, coherent cross-zone relations, and complete scene structure.
In contrast, although HSM and Reason3D obtain slightly better results on a few OP or AO scores, such advantages are largely confined to local placement quality; 
our method is stronger in functional composition, scene richness, and overall structural recovery, which are more critical for fine-grained layout modeling under short-text conditions.

The user ratings in Table~\ref{tab:ablation_compare2} further support this trend. Our method ranks first on all human evaluation criteria, indicating that human judgments likewise favor its advantages in object completeness, region-level organization, and overall coherence.

\begin{table*}[t]
\centering
\caption{Ablation study of our SDesc3D framework under the short-text 3D indoor scene generation setting. MSPA, FLG, and IRR are added progressively to analyze their individual and joint contributions. Each added component consistently improves the results, with the complete model performing best across nearly all metrics.}
\label{tab:ablation}
\vspace{-10pt} 
\setlength{\tabcolsep}{2.5pt}
\footnotesize
\begin{tabular}{c c c c c c c c c c c c c c c c c} 
\toprule
\multirow{2}{*}{\textbf{MSPA}}
& \multirow{2}{*}{\textbf{FLG}}
& \multirow{2}{*}{\textbf{IRR}}
& \multirow{2}{*}{\makecell{\textbf{Collision (\%)}$\downarrow$}} 
& \multirow{2}{*}{\makecell{\textbf{OOB (\%)}$\downarrow$}}
& \multicolumn{2}{c}{\makecell{\textbf{DR} $\uparrow$}} 
& \multicolumn{2}{c}{\makecell{\textbf{OP} $\uparrow$}} 
& \multicolumn{2}{c}{\makecell{\textbf{ZO} $\uparrow$}} 
& \multicolumn{2}{c}{\makecell{\textbf{CR} $\uparrow$}}  
& \multicolumn{2}{c}{\makecell{\textbf{FC} $\uparrow$}} 
& \multicolumn{2}{c}{\textbf{AI-Avg $\uparrow$}} \\
\cmidrule(lr){6-7} \cmidrule(lr){8-9} \cmidrule(lr){10-11} \cmidrule(lr){12-13} \cmidrule(lr){14-15} \cmidrule(lr){16-17}
& & & & 
& \makecell{\scriptsize Gemini\\[-1pt]\scriptsize 3-flash} 
& \makecell{\scriptsize GPT\\[-1pt]\scriptsize 5.4}
& \makecell{\scriptsize Gemini\\[-1pt]\scriptsize 3-flash} 
& \makecell{\scriptsize GPT\\[-1pt]\scriptsize 5.4}
& \makecell{\scriptsize Gemini\\[-1pt]\scriptsize 3-flash} 
& \makecell{\scriptsize GPT\\[-1pt]\scriptsize 5.4}
& \makecell{\scriptsize Gemini\\[-1pt]\scriptsize 3-flash} 
& \makecell{\scriptsize GPT\\[-1pt]\scriptsize 5.4}
& \makecell{\scriptsize Gemini\\[-1pt]\scriptsize 3-flash} 
& \makecell{\scriptsize GPT\\[-1pt]\scriptsize 5.4}
& \makecell{\scriptsize Gemini\\[-1pt]\scriptsize 3-flash} 
& \makecell{\scriptsize GPT\\[-1pt]\scriptsize 5.4} \\
\midrule

& & 
& 38.66\%
& 26.72\%
& 7.43 & 7.26
& 7.41 & 7.02 
& 6.13 & 6.05 
& 6.05 & 5.88 
& 4.78 & 4.56 
& 6.36 & 6.15 \\
\midrule

$\checkmark$
& 
& 
& 19.56\%
& 13.67\%
& 8.48 & 8.30
& 7.85 & 7.42 
& 8.12 & 7.76
& 8.14 & 7.68 
& 8.31 & 7.53
& 8.18 & 7.74 \\
\midrule

$\checkmark$
& $\checkmark$
& 
& 14.28\%
& 8.33\%
& 9.05 & 8.89
& 8.04 & 7.65
& 9.16 & 8.71
& 9.44 & 8.36
& 9.52 & 8.62
& 9.04 & 8.45 \\
\midrule

$\checkmark$
& $\checkmark$
& $\checkmark$
& \textbf{5.44\%}
& \textbf{6.73\%}
& \textbf{9.21} & \textbf{9.03} 
& \textbf{9.25} & \textbf{8.98}
& \textbf{9.54} & \textbf{9.21}
& \textbf{9.53} & \textbf{8.88}
& \textbf{9.88} & \textbf{8.76}
& \textbf{9.48} & \textbf{8.97} \\
\bottomrule
\end{tabular}
\end{table*}

\subsection{Ablation Study}             

We conduct a cumulative ablation over three key components: MSPA, FLG, and IRR. 
\Cref{tab:ablation} shows that these modules play successive roles in short-text scene generation: MSPA mainly compensates for missing scene priors under sparse semantics, FLG further establishes region-level structural organization, and IRR primarily improves physical plausibility and placement precision in the final layouts.

\textbf{MSPA.}
Relative to the base configuration without MSPA, FLG, or IRR, introducing MSPA markedly reduces Collision/OOB from 38.66\%/26.72\% to 19.56\%/13.67\%, while improving AI-Avg from 6.36/6.15 to 8.18/7.74. 
Meanwhile, OP increases from 7.41/7.02 to 7.85/7.42, and FC from 4.78/4.56 to 8.31/7.53; the remaining structural and functional metrics also improve consistently. 
These results indicate that MSPA is necessary for recovering missing scene priors and restoring the basic scene structure under short-text input.

\textbf{FLG.}
Building on MSPA, enabling FLG yields stable performance gains.
Collision/OOB further decrease to 14.28\%/8.33\%, while AI-Avg improves to 9.04/8.45. 
Meanwhile, ZO increases from 8.12/7.76 to 9.16/8.71, and CR from 8.14/7.68 to 9.44/8.36; the remaining structural and functional metrics also continue to improve. 
This shows that FLG provides the critical intermediate structure between high-level scene intent and fine-grained layout recovery, thereby strengthening both region-level organization and cross-zone structural reasoning.

\textbf{IRR.}
Further enabling IRR leads to additional improvements in both physical plausibility and final layout quality. 
Collision/OOB further decrease from 14.28\%/8.33\% to 5.44\%/6.73\%\, while AI-Avg improves from 9.04/8.45 to 9.48/8.97. 
Meanwhile, OP increases from 8.04/7.65 to 9.25/8.98, and FC from 9.52/8.62 to 9.88/8.76; the remaining structural and functional metrics also continue to improve. 
These results show that IRR effectively resolves residual physical conflicts and local instability, while further improving final layout quality.

\textbf{Human-AI score correlation.}
In \Cref{tab:human_human_corr}, both the Pearson $r$ and Spearman $\rho$ between user ratings and AI scores are significantly positive, indicating strong agreement between automatic semantic evaluation and human judgment and thereby supporting the reliability of the automatic analysis.

\begin{table}[t]
\centering
\caption{Correlation Coefficient between User and AI Scores. The results suggest a positive correlation between user judgments and AI-based evaluations.}
\vspace{-10pt}
\label{tab:human_human_corr}
\footnotesize
\setlength{\tabcolsep}{8pt}
\renewcommand{\arraystretch}{1.15}
\begin{tabular}{lccc}
\toprule
\textbf{Metric} & \textbf{User-AI Score Pair 1}  & \textbf{User-AI Score Pair 2}\\
\midrule
Pearson $r$      & 0.81 & 0.76  \\
Spearman $\rho$  & 0.73 & 0.69   \\
\bottomrule
\end{tabular}
\end{table}

\subsection{Comparison on Long-text Conditioned Generation}

\begin{figure}[t]
  \centering
  \includegraphics[width=1.0\linewidth]{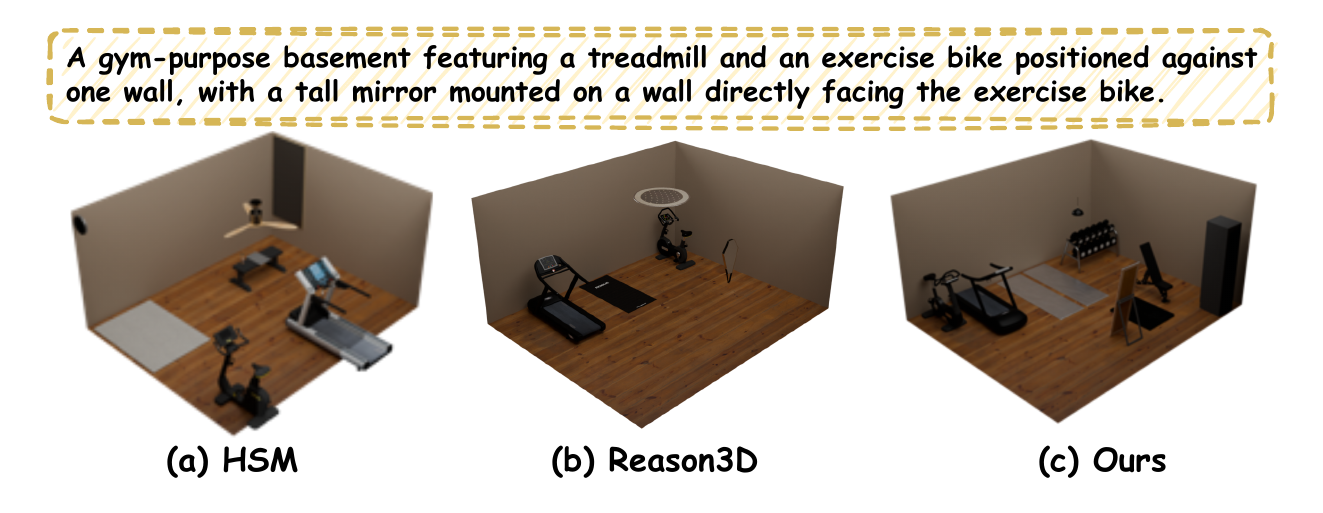}
  \caption{Qualitative comparison of HSM, Reason3D, and our method under the long-text setting.}
  \label{fig:Comparison of Different Methods under Long Texts}
\end{figure}

As shown in \Cref{fig:Comparison of Different Methods under Long Texts}, under long-text inputs, all methods benefit from richer object, attribute, and spatial-relation constraints, leading to more complete indoor scenes.
In this setting, SDesc3D performs comparably to HSM and Reason3D, recovering the major object compositions and local relations described in the text while maintaining clear scene organization and semantic consistency.
These results indicate that although SDesc3D is primarily designed for structural semantic recovery under short-text conditions, its effectiveness is not limited to semantically condensed inputs. 
When richer layout cues are provided, multi-view scene prior augmentation and functionality-aware layout grounding remain effective, suggesting that the framework generalizes well to more descriptive input conditions and continues to support fine-grained, structured scene generation.

\subsection{Analysis on Different LLM Backends.}

We further investigate the performance robustness of SDesc3D across different LLM backends.
As present in \Cref{tab:model_compare}, when the reasoning backend is replaced with GPT-5.4, Gemini 3 Flash, Qwen3, and Claude-sonnet-4-6, the generated scenes remain in a consistently high-performing range under both AI judges with only limited variations, indicating that the effectiveness of the proposed framework does not rely on a single reasoning model and instead generalizes well across different backends.

\begin{table}[t]
\centering
\caption{Model comparison results. The results indicate that our SDesc3D framework exhibits strong robustness across different foundation models.}
\vspace{-10pt} 
\label{tab:model_compare}
\footnotesize
\setlength{\tabcolsep}{10pt}
\begin{tabular}{lcc}
\toprule
\textbf{Model} & \textbf{GPT5.4} & \textbf{Gemini3-flash} \\
\midrule
GPT5.4            &8.72  &8.89  \\
Gemini3-flash     &8.74  &9.06  \\
Qwen3             &8.67  &8.94  \\
Claude-sonnet-4-6 &8.70  &9.03  \\
\bottomrule
\end{tabular}
\end{table}

\subsection{Extending SDesc3D for Indoor Scene Editing}

\begin{figure}[t]  
  \centering
  \includegraphics[width=0.85\linewidth]{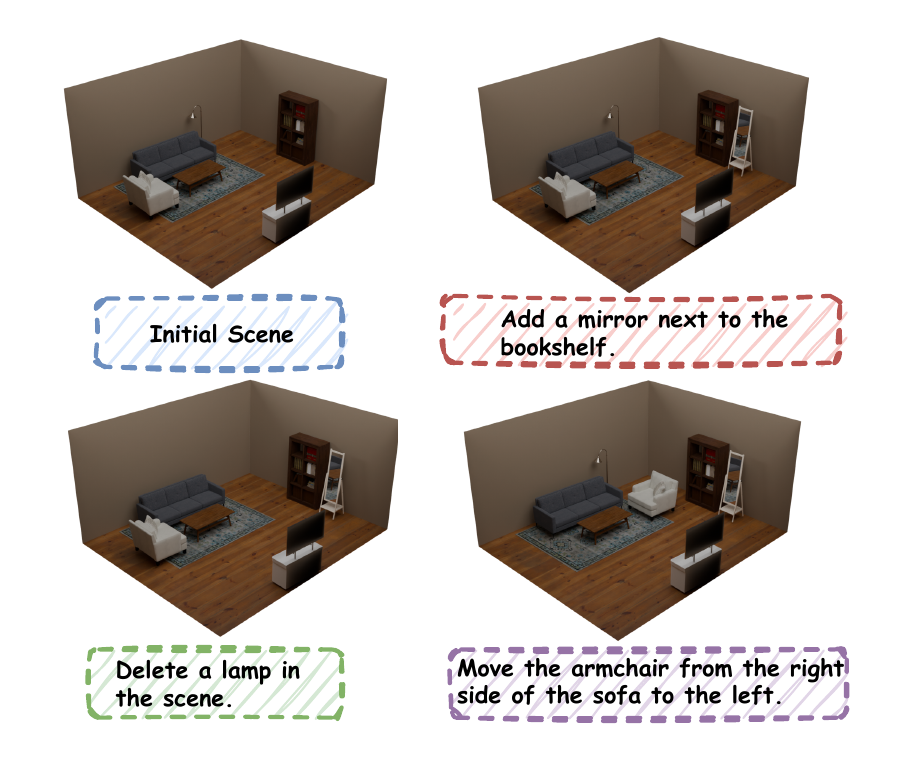}
  \caption{Examples of scene editing results of object addition, deletion and relocation by SDesc3D. Without additional treatments, SDesc3D handles these editing actions with plausible results.}
  \label{fig:editing_task}
  \Description{Editing Task.}
\end{figure}

As shown in Fig.~\ref{fig:editing_task}, SDesc3D also exhibits strong potential for extension to 3D indoor scene editing. 
Enabled by fine-grained functional partitioning and structured scene representation, the framework supports localized editing operations while preserving overall layout consistency and spatial organization stability.
This suggests that SDesc3D does not merely produce a one-shot static layout, but instead captures a fine-grained scene structure with persistent manipulability, thereby providing an extensible representational basis for interactive 3D indoor content creation.

\section{Conclusion}

In our work, we present SDesc3D for indoor layout reasoning conditioned on short scene descriptions. 
To compensate for the lack of explicit semantic guidance on object relationships, our multi-view scene prior augmentation to enrich short descriptions with aggregated multi-view structural knowledge, shifting guidance from inaccessible semantic relation cues to multi-view relational prior aggregation.
Our functionality-aware layout grounding constructs a hierarchical layout reasoning process that leveraging regional functionalities for implicit spatial anchoring, leading to improved scene organization and semantic plausibility.
Experiments validates the superiority of SDesc3D with enhanced physicial plausibility and improved detail richness. 
With its functionality partitioning and hierarchical scene representation, SDesc3D can also be extended for text-based scene editing without additional treatments.


\bibliographystyle{ACM-Reference-Format}
\bibliography{sample-base}

\appendix

\end{document}